\documentclass[sigconf, nonacm]{acmart}
\AtBeginDocument{%
  }

\usepackage{enumitem}
\usepackage{multirow}
\usepackage{booktabs}

\begin{document}

\title{GATE-3D: Geometry-Aware Test-time Adaptive Reranking for Open-Set 3D Shape Retrieval}

\author{Hao Wu}
\affiliation{%
  \institution{Hong Kong University of Science and Technology}
  \city{Hong Kong}
  \country{China}
  }
\affiliation{%
  \institution{Hong Kong University of Science and Technology (Guangzhou)}
  \city{Guangzhou}
  \country{China}
  }

\author{Heyi Lin}
\affiliation{%
  \institution{Hong Kong University of Science and Technology (Guangzhou)}
  \city{Guangzhou}
  \country{China}
  }

\author{Zilin Wang}
\affiliation{%
  \institution{Hong Kong University of Science and Technology (Guangzhou)}
  \city{Guangzhou}
  \country{China}
  }

\author{Huizai Yao}
\affiliation{%
  \institution{Hong Kong University of Science and Technology (Guangzhou)}
  \city{Guangzhou}
  \country{China}
  }

\author{Hao Wang}
\affiliation{%
  \institution{Hong Kong University of Science and Technology (Guangzhou)}
  \city{Guangzhou}
  \country{China}
  }

\author{Hui Xiong}
\authornote{Corresponding author.}
\affiliation{%
  \institution{Hong Kong University of Science and Technology (Guangzhou)}
  \city{Guangzhou}
  \country{China}
  }
\affiliation{%
  \institution{Hong Kong University of Science and Technology}
  \city{Hong Kong}
  \country{China}
  }
\renewcommand{\shortauthors}{Wu et al.}

\begin{abstract}
Large pretrained vision models have substantially improved appea\-rance-based 3D shape retrieval, but they still confuse shapes that look similar while differing in geometry. Although geometry-aware features can reduce these errors, naive fusion of geometry and appearance may hurt retrieval when the two modalities are already well aligned. We propose \textbf{GATE-3D}, a lightweight query-adaptive reranking method that incorporates geometry without retraining the retrieval backbone. For each query, GATE-3D predicts how much a geometry-aware score should adjust the appearance-based ranking using features that capture disagreement between the two modalities. This selective design lets geometry contribute where it helps and stay silent where it would hurt. Experiments on three open-set 3D retrieval benchmarks show that GATE-3D improves over appearance-only retrieval and is more robust than always-on fusion. On the primary benchmark, it improves mAP@10 by 2.00 points over appearance-only retrieval (p=0.041); it also improves leave-one-category-out generalization and reduces geometric false positives by 10.8\%. GATE-3D achieves competitive zero-shot results against DAC-based baselines. We further find that simple linear routing is more effective than a small MLP in the low-data regime, suggesting that cross-modal disagreement features matter more than model capacity for adaptive routing.
\end{abstract}

\begin{CCSXML}
<ccs2012>
   <concept>
       <concept_id>10010147.10010178.10010224.10010225.10010231</concept_id>
       <concept_desc>Computing methodologies~Visual content-based indexing and retrieval</concept_desc>
       <concept_significance>500</concept_significance>
       </concept>
   <concept>
       <concept_id>10010147.10010178.10010224.10010240.10010242</concept_id>
       <concept_desc>Computing methodologies~Shape representations</concept_desc>
       <concept_significance>300</concept_significance>
       </concept>
   <concept>
       <concept_id>10010147.10010178.10010224.10010240.10010243</concept_id>
       <concept_desc>Computing methodologies~Appearance and texture representations</concept_desc>
       <concept_significance>100</concept_significance>
       </concept>
 </ccs2012>
\end{CCSXML}

\ccsdesc[500]{Computing methodologies~Visual content-based indexing and retrieval}
\ccsdesc[300]{Computing methodologies~Shape representations}
\ccsdesc[100]{Computing methodologies~Appearance and texture representations}

\keywords{3D shape retrieval, geometry-aware features, test-time adaptation, cross-modal routing}

\maketitle

%% ====================================================================
%%  1. INTRODUCTION
%% ====================================================================
\section{Introduction}
\label{sec:intro}

Three-dimensional shape retrieval is foundational to engineering applications such as computer-aided design, manufacturing, and digital asset management~\cite{jayanti2006esb, chen2003ntu3d}.
Given a query 3D shape, the goal is to retrieve geometrically and semantically similar shapes from a large gallery---a capability that directly supports part reuse, quality inspection, and design search in industrial workflows.
The \emph{open-set} setting, where test-time queries may belong to categories absent during training, makes this problem particularly challenging and practically relevant~\cite{feng2023hypergraph}.

Recent advances in large pretrained vision models have substantially improved appearance-based 3D retrieval.~\cite{bai2016gift, he2018triplet, hamdi2021mvtn}
Self-supervised models such as DINOv2~\cite{oquab2023dinov2} produce rich spatial features from multi-view renderings without domain-specific training, achieving strong performance on general shape benchmarks.
Vision-language models like CLIP~\cite{radford2021clip} provide complementary semantic priors through language-aligned visual representations.
Building on these foundations, DAC~\cite{wang2025dac} combines CLIP adaptation via AB-LoRA with textual descriptors generated by multimodal LLMs, achieving state-of-the-art results on open-set benchmarks.
HGM$^2$R~\cite{feng2023hypergraph} further leverages hypergraph-based high-order correlation modeling to learn unified multi-modal 3D object embeddings.
Zero-shot text-3D pretraining methods---OpenShape~\cite{liu2023openshape}, ULIP-2~\cite{xue2024ulip}, and Uni3D~\cite{zhou2023uni3d}---align point cloud encoders with CLIP features, enabling retrieval without any target-domain supervision.~\cite{yu2022point, zhang2022point, zhang2023learning, zhang2022pointclip}

\begin{figure*}[t]
  \centering
  \includegraphics[width=0.78\textwidth]{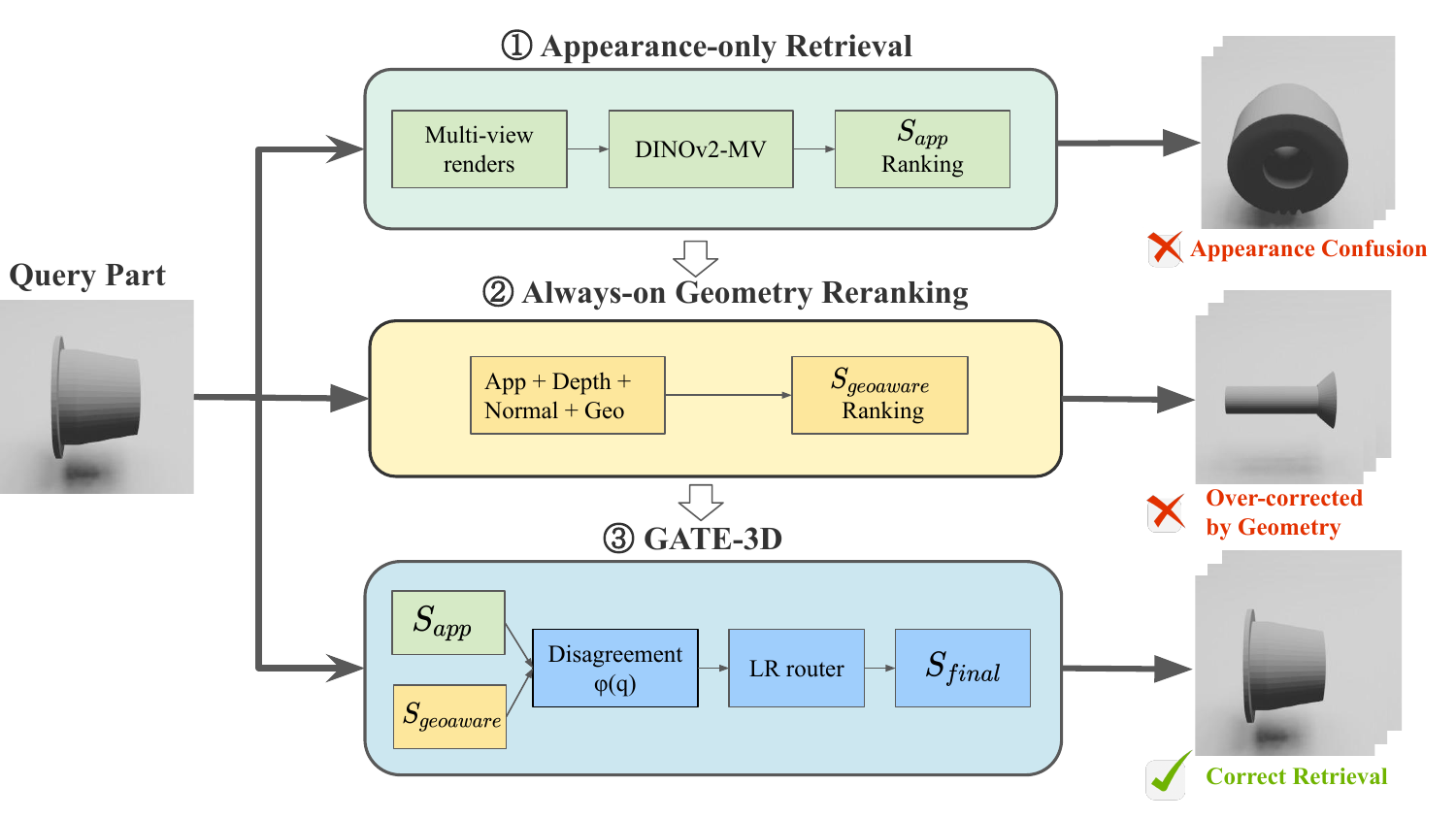}
  \caption{Motivation for query-adaptive geometry routing.
  Given a mechanical query part,
  \textcircled{1}~appearance-only retrieval (DINOv2-MV) returns a visually similar
  but geometrically incorrect result (appearance confusion);
  \textcircled{2}~always-on geometry reranking over-corrects the ranking
  by applying geometry indiscriminately;
  \textcircled{3}~GATE-3D selectively activates geometry based on
  cross-modal disagreement, producing the correct retrieval.
  This illustrates that geometry is not uniformly helpful
  and motivates per-query adaptive routing.}
  \Description{Three-row comparison showing appearance-only retrieval
  producing a geometric false positive, always-on fusion over-correcting,
  and GATE-3D correctly retrieving the target shape via selective routing.}
  \label{fig:motivation}
\end{figure*}

Despite their effectiveness on general consumer-object benchmarks, these methods share a common limitation: \emph{they rely predominantly on appearance-level or semantic-level representations while lacking explicit geometric reasoning}.
CLIP-based methods capture object-level semantics but are largely invariant to fine-grained structural differences.
DINOv2 encodes richer spatial information, yet remains a 2D feature extractor operating on rendered views.
For geometry-sensitive domains---particularly mechanical part retrieval---this gap is critical.~\cite{fu2020risa}
Shapes that appear visually similar in rendered views can differ significantly in structure: a blind hole versus a through-hole, a flat flange versus a conical one.
We call such ranking errors \emph{geometric false positives} (GFPs): gallery items that score highly under appearance-based similarity despite being geometrically dissimilar to the query.
On the open-set mechanical part benchmark OS-ESB-core~\cite{feng2023hypergraph}, even the strong DINOv2 baseline exhibits a GFP rate of 12.3\% at rank~10---a level that can propagate errors into downstream CAD workflows. 
Figure \ref{fig:motivation} illustrates this failure pattern on a concrete example: appearance-only retrieval confuses a flange with a geometrically dissimilar part, while always-on geometry fusion over-corrects the ranking. This motivates a selective mechanism that activates geometry only when the two modalities disagree.

A natural remedy is to incorporate geometry-aware features extracted from depth maps, surface normals, or point clouds.
However, doing so is non-trivial.
Multi-modal fusion methods such as MV-MAE~\cite{chen2025mae} and HGM$^2$R combine modalities at training time in a fixed manner, requiring end-to-end retraining whenever the feature set changes.
More fundamentally, we find that naively fusing geometry with appearance at retrieval time (``always-on'' fusion) yields only modest improvements, because geometry is \emph{not uniformly informative}: it helps some queries but introduces noise for others.
An oracle that activates geometry-aware reranking only when it actually improves retrieval achieves a substantially larger gain than unconditional fusion, revealing a significant gap left on the table.
This observation motivates a central question: \emph{can we predict, at test time, when and how much geometry should contribute to retrieval for a given query?}

We address this question with \textbf{GATE-3D} (\textbf{G}eometry-\textbf{A}ware \textbf{T}est-time r\textbf{E}ranker for \textbf{3D} shapes), a lightweight, query-adaptive reranking framework that requires no backbone retraining.
Our key insight is that \emph{cross-modal disagreement}---the degree to which appearance-based and geometry-based rankings diverge over retrieved candidates---provides a strong signal for deciding per-query geometry activation.
Specifically, GATE-3D first retrieves candidates using appearance features, then computes a compact set of disagreement features characterizing how the appearance and geometry rankings differ over these candidates.
These features are used to predict a per-query blending weight that controls how much of a geometry residual correction is added to the appearance score.
When the predicted weight is zero, the system defaults to pure appearance-based retrieval; when it is one, full geometry-aware fusion is applied; and intermediate values enable graded correction for ambiguous queries.

We instantiate GATE-3D in two forms: a continuous variant (GATE-3D-$\alpha$) based on Ridge regression that predicts a real-valued blending weight, and a binary variant based on logistic regression that makes a hard routing decision.
Both are trained under a strict out-of-fold cross-validation protocol, and both operate as a drop-in reranking stage on top of \emph{any} appearance-based retrieval backbone.
A notable empirical finding is that simple linear models substantially outperform MLPs in this low-data routing regime, underscoring that the quality of cross-modal disagreement features matters more than model capacity.

We evaluate GATE-3D on three open-set 3D retrieval benchmarks.
On the geometry-sensitive OS-ESB-core benchmark, GATE-3D achieves statistically significant improvements over the appearance-only baseline, reduces geometric false positives, and generalizes to unseen shape categories under leave-one-category-out evaluation.
On general-object benchmarks (OS-MN40-core, OS-NTU-core)~\cite{feng2023hypergraph} where geometry is less discriminative, GATE-3D correctly defaults to near-appearance-only behavior, avoiding the degradation observed with always-on fusion.
Without any target-domain fine-tuning, our zero-shot pipeline achieves competitive results against supervised baselines including DAC~\cite{wang2025dac}.

Our contributions are summarized as follows:
\begin{itemize}[leftmargin=*]
  \item We propose GATE-3D, a query-adaptive geometry reranking framework that formulates geometry activation as a residual correction to appearance-based retrieval. The framework is modular, backbone-agnostic, and requires no retraining of the underlying feature extractors.
  \item We design a 24-dimensional cross-modal disagreement feature set that captures rank correlation, score distribution, and margin divergence between appearance and geometry modalities, with grouped ablations confirming the complementarity of different feature groups.
  \item We conduct rigorous evaluation under out-of-fold and leave-one-category-out protocols across three benchmarks, demonstrating that selective routing is more robust than unconditional fusion, particularly on geometry-sensitive domains and in cross-dataset transfer scenarios.
\end{itemize}

%% ====================================================================
%%  2. RELATED WORK
%% ====================================================================
\section{Related Work}
\label{sec:related}

\paragraph{Open-Set 3D Shape Retrieval.}
Open-set retrieval requires models to generalize to shape categories that are absent during training.
Classical view-based 3D retrieval systems such as GIFT and metric-learning approaches such as Triplet-Center Loss established strong baselines for multi-view shape retrieval, while later work such as MVTN showed that view selection itself can be learned rather than fixed.
Existing work broadly follows two directions.
One line adapts vision-language models to 3D retrieval, as in DAC~\cite{wang2025dac}, which extends CLIP-based retrieval with textual descriptors and AB-LoRA fine-tuning, and TeDA~\cite{wang2025teda}, which studies test-time distribution alignment.
Another line emphasizes structural modeling across modalities.
HGM$^2$R~\cite{feng2023hypergraph}, HARF~\cite{xu2024harf}, and HERT~\cite{xu2024hert} use hypergraph-based formulations to capture higher-order, relational, or hierarchical structure under open-set or semi-open settings.
Related zero-shot text--3D pre-training methods, including OpenShape~\cite{liu2023openshape}, ULIP-2~\cite{xue2024ulip}, and Uni3D~\cite{zhou2023uni3d}, align 3D encoders with vision-language embedding spaces to improve category-level generalization; related 3D representation transfer and pretraining approaches include Point-BERT, Point-M2AE, I2P-MAE, and PointCLIP.~\cite{yu2022point, zhang2022point, zhang2023learning, zhang2022pointclip}

\paragraph{Multi-Modal 3D Feature Representations.}
Different 3D modalities capture complementary cues.
Depth maps, surface normals, and point clouds preserve geometric information that is often only implicit in RGB renderings.
Multi-view CNNs~\cite{su2015multi} were among the earliest successful approaches based on rendered views, while PointNet~\cite{qi2017pointnet} established direct point-cloud processing.
Subsequent work~\cite{wang2022p2p, zhu2023pointclip} explored aligning 3D representations with 2D vision-language models.
CLIP-based multi-view representations~\cite{radford2021clip} are effective baselines for semantic retrieval, but they do not directly model fine-grained 3D geometry.
Methods such as MV-MAE~\cite{chen2025mae} instead learn joint 3D and multi-view representations within a masked multi-modal framework.

\paragraph{Test-Time Adaptation and Query Routing.}
Test-time training~\cite{sun2020test} and test-time adaptation methods~\cite{liang2025comprehensive} modify model behavior during inference, typically by updating model parameters with auxiliary objectives.
Our setting is closer to retrieval reranking~\cite{zhong2017re}, since the feature extractor is fixed and the adaptation occurs at the scoring stage.
Related ideas also appear in dynamic multimodal fusion frameworks, where gating is used to generate input-dependent fusion paths or select dominant modalities~\cite{xue2023dynamic,guo2024damsdet}.
These methods, however, are usually trained end-to-end for the target architecture and task.
In contrast, GATE-3D uses a lightweight gate only to adjust retrieval scores at test time over a fixed backbone. Our routing mechanism is also related to adaptive multimodal fusion, where gated units and dynamic late-fusion methods make input-dependent decisions about how much each modality should contribute.
We use the term test-time adaptation in a score-level sense: unlike classical TTA methods that update model parameters on target data, GATE-3D keeps all feature extractors frozen and adapts only the query-time score blending weight.
%% =====================================================================
%%  SECTION 3 — METHOD
%% =====================================================================
\section{Method}
\label{sec:method}

\begin{figure*}[t]
  \centering
  \includegraphics[width=0.82\textwidth]{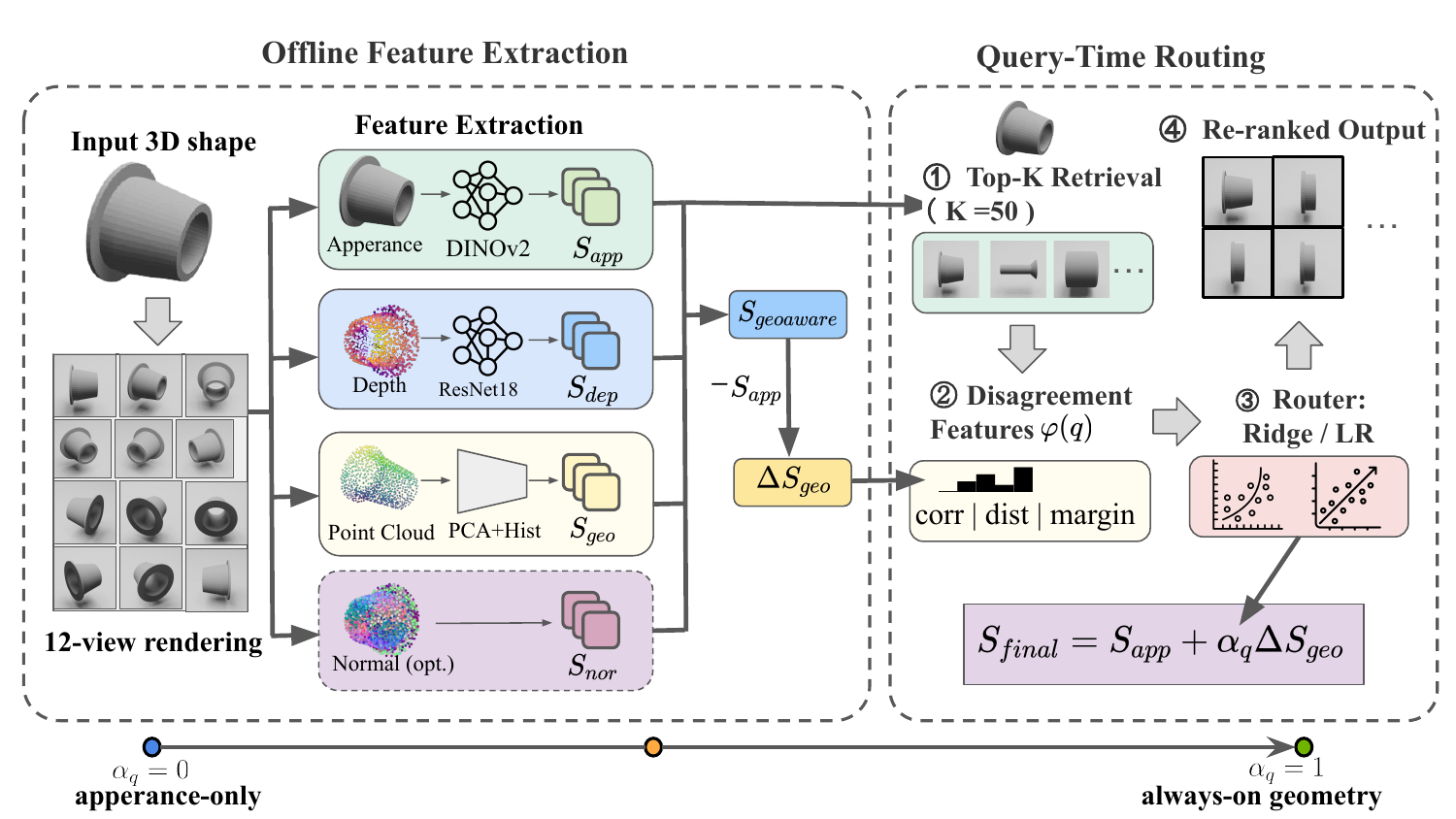}
  \caption{Overview of the GATE-3D framework.
  \textit{Left} (Offline Feature Extraction): each 3D shape is rendered
  into 12 views and encoded by DINOv2 (appearance), a randomly
  initialized ResNet18 (depth), and a PCA+histogram descriptor
  (point cloud); surface normals are optionally included.
  A geometry residual \(\Delta S_{\text{geo}}\) is computed as the
  difference between the geometry-aware aggregate
  \(S_{\text{geoaware}}\) and the appearance score \(S_{\text{app}}\).
  \textit{Right} (Query-Time Routing): for each query, top-\(K\)
  candidates are retrieved using \(S_{\text{app}}\); cross-modal
  disagreement features \(\varphi(\mathbf{q})\) are extracted; a
  lightweight router predicts a per-query blending weight
  \(\alpha_q \in [0,1]\), yielding the final score
  \(S_{\text{final}} = S_{\text{app}} + \alpha_q \cdot \Delta S_{\text{geo}}\).
  The slider at bottom illustrates that \(\alpha_q = 0\) defaults to
  appearance-only retrieval and \(\alpha_q = 1\) applies full
  geometry fusion.}
  \Description{Two-part pipeline diagram showing offline multi-modal
  feature extraction on the left and query-time adaptive routing
  with disagreement features and a linear router on the right.}
  \label{fig:pipeline}
\end{figure*}

Figure~\ref{fig:pipeline} provides an overview of GATE-3D.
Given a query 3D shape, we first retrieve top-$K$ candidates using appearance features from a pretrained vision backbone.
We then compute a geometry-aware score for each candidate by combining depth, point cloud, and (optionally) surface normal similarities.
The key idea is to predict a per-query blending weight $\alpha_q \in [0,1]$ from \emph{cross-modal disagreement features}---statistics that measure how much the appearance and geometry rankings diverge---and use it to control how much geometry correction is applied to the appearance-based ranking.
This design allows GATE-3D to activate geometry only where it helps and remain silent where it would introduce noise.
Crucially, the entire framework operates as a lightweight post-hoc reranking stage and requires no retraining of the underlying feature extractors. 
All gallery features are pre-extracted offline; at query time, only the query's features require computation. Given these multi-modal scores, the central question is how to decide \emph{when} geometry should override appearance---we address this next.

\subsection{Preliminaries: Multi-Modal Scoring}
\label{sec:feature_extraction}

\paragraph{Appearance score.}
For each 3D shape, we render 12 views under uniform lighting and extract DINOv2-ViT-B/14~\cite{oquab2023dinov2} features per view, aggregated via max-pooling into a 768-dimensional descriptor $\mathbf{f}_{\text{app}}$.
The appearance similarity between query $\mathbf{q}$ and gallery item $i$ is:
\begin{equation}
    S_{\text{app}}(\mathbf{q}, i) = \cos(\mathbf{f}^{\mathbf{q}}_{\text{app}},\; \mathbf{f}^{i}_{\text{app}}).
    \label{eq:sapp}
\end{equation}

\paragraph{Geometry scores.}
We extract two complementary geometry representations.
\emph{Depth features}: depth maps rendered under the same 12 viewpoints are encoded by a randomly initialized ResNet18~\cite{he2016resnet} (512-dim, max-pooled), yielding depth similarity $S_{\text{dep}}(\mathbf{q}, i)$.
We use random initialization rather than ImageNet pretraining to capture low-level geometric structure (depth gradients, surface topology) without interference from object-level semantics; this yields a +5.9\,pp mAP improvement on OS-ESB-core.
\emph{Point cloud features}: we sample 2048 surface points per shape and compute a PCA\,+\,histogram descriptor (33-dim) capturing spatial extent and local curvature~\cite{zhou2018open3d}, yielding $S_{\text{geo}}(\mathbf{q}, i)$.
Surface normal maps are retained as an optional modality but contribute negligibly ($\alpha_{\text{nor}} = 0.014$; see Section~\ref{sec:ablation}).

\paragraph{Geometry-aware aggregate and residual.}
We combine all modality scores into a geometry-aware aggregate via calibrated weights:
\begin{equation}
    S_{\text{geoaware}}(\mathbf{q}, i) = \alpha_{\text{app}}\, S_{\text{app}} + \alpha_{\text{dep}}\, S_{\text{dep}} + \alpha_{\text{nor}}\, S_{\text{nor}} + \alpha_{\text{geo}}\, S_{\text{geo}},
    \label{eq:geoaware}
\end{equation}
where the weights $(\alpha_{\text{app}}, \alpha_{\text{dep}}, \alpha_{\text{nor}}, \alpha_{\text{geo}})$ are determined by Dirichlet-sampled nested grid search on training queries under nested cross-validation.
Central to our formulation is the \emph{geometry residual}:
\begin{equation}
    \Delta S_{\text{geo}}(\mathbf{q}, i) = S_{\text{geoaware}}(\mathbf{q}, i) - S_{\text{app}}(\mathbf{q}, i),
    \label{eq:residual}
\end{equation}
which isolates the pure geometric correction relative to appearance.
When the two modalities agree, $\Delta S_{\text{geo}} \approx 0$; when geometry would promote a gallery item beyond what appearance predicts, $\Delta S_{\text{geo}} > 0$.
All gallery features are pre-extracted offline; at query time, only the query's features require computation.

\subsection{Cross-Modal Disagreement Features}
\label{sec:disagreement}

The routing decision in GATE-3D is driven by a 24-dimensional feature vector $\boldsymbol{\varphi}(\mathbf{q})$ that characterizes how strongly the appearance and geometry rankings diverge for a given query.
After retrieving the top-$K$ candidates ($K{=}50$) using $S_{\text{app}}$ and computing $S_{\text{geoaware}}$ for each, we extract three groups of statistics:

\paragraph{Rank correlations (8 features).}
Kendall-$\tau$~\cite{kendall1938new} and Spearman-$\rho$~\cite{spearman1987proof} between the appearance ranking and each geometry modality ranking (depth, normal, point cloud), as well as the combined $S_{\text{geoaware}}$ ranking.
These capture the overall alignment between modalities at the ranking level.

\paragraph{Score distribution (8 features).}
Mean absolute deviation of score differences across the $K$ candidates per modality pair, and the fraction of top-10 candidates exhibiting rank inversions between geometry and appearance.
These capture the magnitude and localization of disagreement.

\paragraph{Margin features (8 features).}
The score gap between top-1 and top-5 candidates under each modality, capturing how peaked each retrieval distribution is.
A flat distribution indicates retrieval ambiguity, which often correlates with potential geometry benefit.

Among individual features, the Kendall-$\tau$ between appearance and the combined geometry-aware ranking ($\tau_{\text{app-geo}}$) is most predictive (Pearson $r{=}{-0.245}$ with $\Delta$AP, $p{=}0.007$), but no single feature is sufficient; the full 24-dimensional set achieves the best out-of-fold (OOF) performance, and grouped ablations confirm inter-group complementarity (Section~\ref{sec:ablation}).

\subsection{Query-Adaptive Geometry Routing}
\label{sec:routing}

Given the disagreement features $\boldsymbol{\varphi}(\mathbf{q})$, GATE-3D predicts a per-query activation weight $\alpha_q$ and produces the final retrieval score:
\begin{equation}
    S_{\text{final}}(\mathbf{q}, i) = S_{\text{app}}(\mathbf{q}, i) + \alpha_q \cdot \Delta S_{\text{geo}}(\mathbf{q}, i).
    \label{eq:sfinal}
\end{equation}
This residual formulation provides clean routing semantics: $\alpha_q{=}0$ defaults to pure appearance retrieval; $\alpha_q{=}1$ applies full geometry-aware fusion; and intermediate values enable graded correction.
We implement two instantiations:

\paragraph{GATE-3D-$\boldsymbol{\alpha}$ (continuous routing).}
Ridge regression maps $\boldsymbol{\varphi}(\mathbf{q})$ to a continuous activation weight:
\begin{equation}
    \hat{\alpha}_q = \text{clip}\!\left(\mathbf{v}^{\top} \boldsymbol{\varphi}(\mathbf{q}) + c,\; 0,\; 1\right),
    \label{eq:alpha}
\end{equation}
where L2 regularization strength $\lambda$ is tuned by inner 3-fold CV.
The regression target is the per-query oracle alpha:
\begin{equation}
    \alpha^{*}_q = \arg\max_{\alpha \in \{0, 0.1, \ldots, 1.0\}} \text{AP}\!\left(S_{\text{app}} + \alpha \cdot \Delta S_{\text{geo}}\right),
    \label{eq:oracle_alpha}
\end{equation}
obtained by grid search over the candidate pool.
We verify that finer grids ($\Delta\alpha \in \{0.05, 0.1, 0.2\}$) yield negligible variation ($<$0.1\,pp OOF mAP).
A conservative fallback clips $\hat{\alpha}_q \to 0$ when $|\hat{\alpha}_q| < \epsilon$ (default $\epsilon{=}0.15$), complementing the implicit shrinkage from L2 regularization.

\paragraph{GATE-3D LR (binary routing).}
As a simplified variant, a logistic regression gate $g(\mathbf{q}) = \sigma(\mathbf{w}^{\top}\boldsymbol{\varphi}(\mathbf{q}) + b)$ produces a binary decision:
\begin{equation}
    \alpha^{\text{binary}}_q = \mathbf{1}[g(\mathbf{q}) > 0.5].
    \label{eq:binary}
\end{equation}
Binary routing is effective when geometry clearly helps or hurts on a per-query basis; continuous routing is preferable when many queries benefit from partial geometry activation.

\subsection{Why Linear Models Suffice}
\label{sec:model_choice}

With only 120 training queries and 22 positives (18.3\% positive rate), the routing problem is severely data-limited.
An MLP with 2 hidden layers (64 units, 1345 parameters) memorizes training data and produces near-zero OOF gate activation---a textbook overfitting failure.
In contrast, logistic regression with L2 regularization (25 parameters, $C{=}0.1$) provides an appropriate bias-variance tradeoff.
The L2 penalty implicitly pushes uncertain predictions toward the null action (appearance-only retrieval), making the gate conservative by construction.
Ablation confirms the gap: LR achieves +2.00\,pp OOF and +1.56\,pp LOCV, while MLP achieves only +0.76\,pp OOF and +0.00\,pp LOCV (Section~\ref{sec:ablation}).
This finding suggests that, in the low-data routing regime, the quality of input features matters more than model capacity.

\subsection{Training Protocol}
\label{sec:protocol}

GATE-3D is trained and evaluated under a strict \textbf{out-of-fold (OOF)} protocol: 5-fold cross-validation repeated over 5 random seeds.
The gate is trained on 4 folds and evaluated on the held-out fold; all hyperparameters, feature standardization statistics, and geometry-fusion weights are estimated exclusively from training-fold queries.
For open-set generalization, we additionally evaluate under \textbf{Leave-One-Category-Out CV (LOCV)}: each of the 24 fine-grained subcategories in OS-ESB-core is held out entirely (5 queries per category), directly testing whether the routing signal transfers to unseen shape categories.
We verify candidate pool sensitivity across $K \in \{20, 30, 50, 100\}$: $K{=}50$ yields the best inner-fold mAP@10; at extreme $K$ values, $\hat{\alpha}_q$ shrinks toward zero, providing a safe fallback.

All modality scores are cosine similarities in $[-1, 1]$.
The feature vector $\boldsymbol{\varphi}(\mathbf{q})$ is z-score standardized per outer fold using training-query statistics only.
On OS-NTU-core, where depth/normal renders are unavailable, we substitute ResNet18-MV (random-init, 12 views, max-pooled) for $S_{\text{dep}}$ and $S_{\text{nor}}$, and PointNet (1024 points) for $S_{\text{geo}}$ as proxy modalities.

\subsection{Backbone Flexibility and Extensions}
\label{sec:extensions}

GATE-3D is agnostic to the choice of appearance backbone.
In addition to DINOv2, we support CLIP-ViT-L/14~\cite{radford2021clip} as an alternative, whose language-aligned visual features benefit semantically diverse benchmarks.
For maximum generalization, the two backbones can be combined via a convex score-level ensemble:
\begin{equation}
    S_{\text{ens}}(\mathbf{q}, i) = (1 - w)\, S_{\text{DINOv2}}(\mathbf{q}, i) + w\, S_{\text{CLIP}}(\mathbf{q}, i),
    \label{eq:ensemble}
\end{equation}
where $w \in [0,1]$ is selected by inner-fold cross-validation.
We also employ Query Expansion (QE)~\cite{zhong2017re} as an optional post-processing step, which refines the query descriptor by averaging with top-$K$ gallery features.
On OS-NTU-core, QE yields +1.45\,pp mAP at negligible computational cost.
%% =====================================================================
%%  SECTION 4 — EXPERIMENTS
%% =====================================================================
\section{Experiments}
\label{sec:experiments}

\subsection{Setup}
\label{sec:setup}

\paragraph{Datasets.}
We evaluate on three open-set 3D shape retrieval benchmarks following the DAC~\cite{wang2025dac} protocol. 
These benchmarks, namely OS-ESB-core, OS-NTU-core, and OS-MN40-core, were introduced in HGM2R~\cite{feng2023hypergraph} and are derived from the Engineering Shape Benchmark~\cite{jayanti2006esb}, NTU 3D Model Benchmark~\cite{chen2003ntu3d}, and ModelNet40 datasets~\cite{wu20153modelnet}, respectively.
\textbf{OS-ESB-core} contains 120 queries and 452 gallery shapes across 24 fine-grained mechanical part subcategories (e.g., cylinders, brackets, flanges), with disjoint categories between train and test splits. This is our primary benchmark where geometry is most discriminative.
\textbf{OS-MN40-core} contains 160 queries and 9,482 gallery shapes spanning general consumer objects (e.g., chairs, tables, lamps), serving as a domain where appearance already captures most discriminative structure.
\textbf{OS-NTU-core} contains 270 queries and 1,271 gallery shapes across diverse categories. As depth/normal renders are unavailable in the distributed split, we use proxy geometry modalities (ResNet18-MV for depth/normal, PointNet for point cloud features).

\paragraph{Metrics.}
We report three complementary metrics: \textbf{standard mAP} (normalized by total relevant items $G$, rewarding complete recall), \textbf{NDCG} (following the DAC evaluation protocol with NDCG@100 and fixed IDCG denominator for cross-method comparability), and \textbf{ANMRR}~\cite{cieplinski2001mpeg} (MPEG-7 formula, lower is better).
We additionally report mAP@10 on OS-ESB-core for statistical testing of the OOF improvement.

\paragraph{Baselines.}
We compare against three groups of methods:
(i)~\emph{Zero-shot text-3D encoders} from DAC~\cite{wang2025dac}: OpenShape~\cite{liu2023openshape}, ULIP-2~\cite{xue2024ulip}, Uni3D~\cite{zhou2023uni3d}, and DAC zero-shot;
(ii)~\emph{Multi-view image and 3D baselines} (our computation): ResNet18-MV~\cite{he2016resnet}, PointNet~\cite{qi2017pointnet}, VoxNet~\cite{maturana2015voxnet}, MV-MAE~\cite{chen2025mae}, DINOv2-MV~\cite{oquab2023dinov2}, and CLIP-MV~\cite{radford2021clip};
(iii)~\emph{Supervised open-set methods}: HGM$^2$R~\cite{feng2023hypergraph} and DAC with AB-LoRA~\cite{wang2025dac}.
All multi-view methods use identical rendered images. Our routing follows the strict OOF protocol described in Section~\ref{sec:protocol}.

\begin{figure*}[t]
  \centering
  \includegraphics[width=0.78\textwidth]{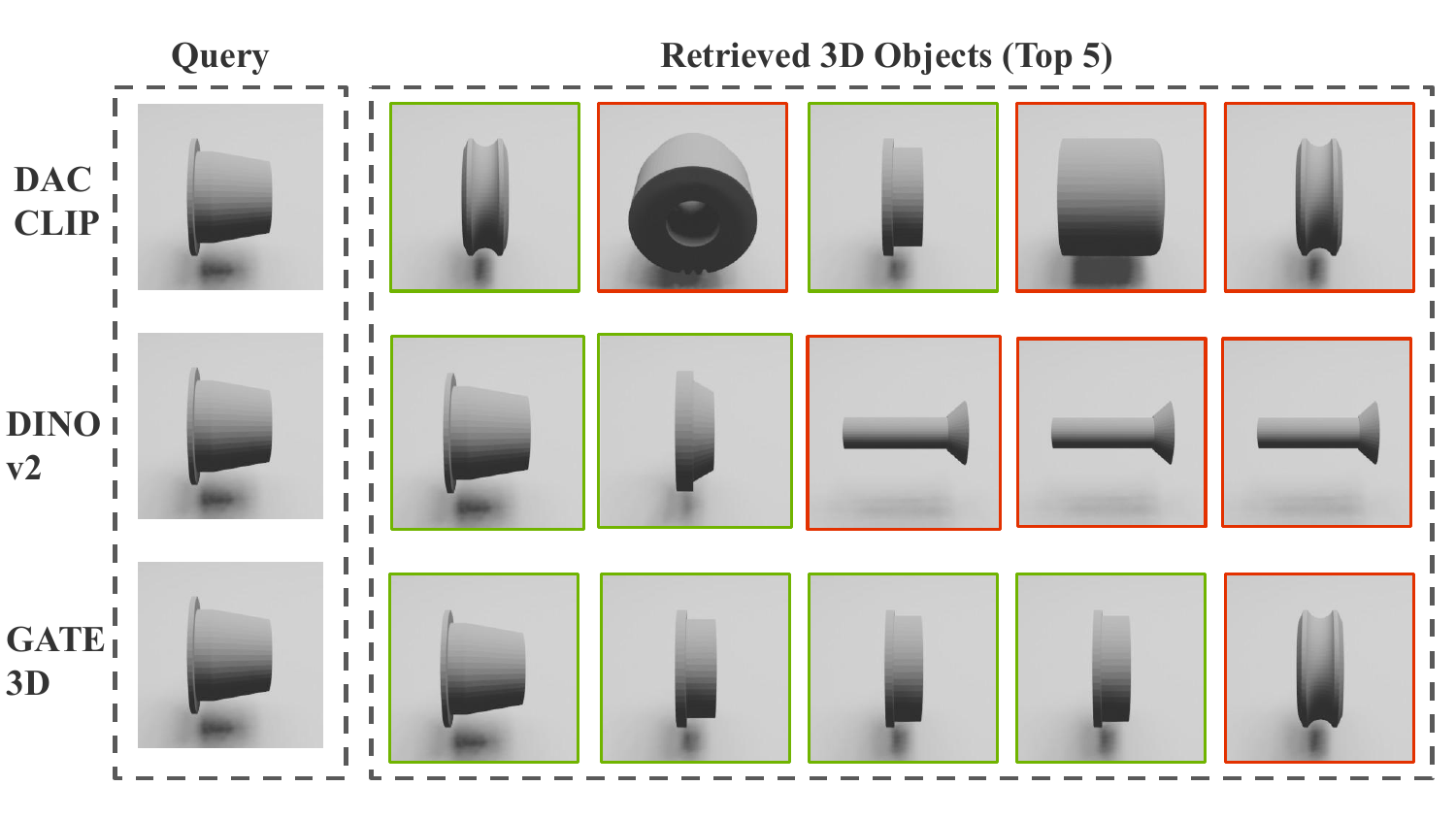}
  \caption{Qualitative retrieval comparison on OS-ESB-core.
  Top-5 retrieved shapes for a flange-type query under
  DAC (CLIP), DINOv2-MV, and GATE-3D.
  Green borders denote correct retrievals (same subcategory);
  red borders denote incorrect ones.
  DAC and DINOv2 each return 3 incorrect results due to
  appearance-level confusion among visually similar mechanical parts.
  GATE-3D correctly retrieves 4 out of 5 by leveraging geometry
  where cross-modal disagreement signals its benefit.}
  \Description{Grid of top-5 retrieval results for three methods,
  showing GATE-3D achieving four correct retrievals compared
  to two each for DAC and DINOv2.}
  \label{fig:qualitative}
\end{figure*}

\subsection{Main Results}
\label{sec:main_results}

Table~\ref{tab:main} presents results across all three benchmarks.

\paragraph{OS-ESB-core (geometry-sensitive).}
DINOv2-MV is the strongest single-encoder zero-shot baseline (mAP 57.02, ANMRR 38.70).
Always-on geometry fusion improves mAP to 58.65 (+1.63\,pp), with ANMRR\,=\,36.87---10.6\,pp better than supervised DAC (47.44).
GATE-3D-$\alpha$ matches Always-on on standard mAP (58.58 vs.\ 58.65) while using adaptive routing.
On mAP@10, GATE-3D LR achieves +2.00\,pp over DINOv2 ($p{=}0.041$, one-sided paired $t$-test)---the only statistically significant improvement among all methods.
Notably, our zero-shot geometry-aware pipeline surpasses both DAC zero-shot (56.60) and DAC supervised (57.80) on standard mAP without any target-domain fine-tuning.
Figure \ref{fig:qualitative} provides a qualitative example: GATE-3D correctly retrieves 4/5 flange parts where both DAC and DINOv2 return only 2/5.

\paragraph{OS-MN40-core (appearance-dominated).}
On this benchmark, appearance already captures most discriminative structure: App.\ Fusion (DINOv2+CLIP) achieves the best zero-shot result (mAP 63.64, ANMRR 29.74).
Always-on geometry fusion slightly degrades performance (62.15 vs.\ DINOv2's 62.26), confirming that geometry introduces noise on general consumer objects.
Both GATE-3D variants recover from this degradation: GATE-3D-$\alpha$ achieves 62.66 (+0.51\,pp over Always-on), demonstrating the safety advantage of selective routing.
Both variants also outperform supervised DAC on ANMRR (30.65 and 30.80 vs.\ 33.87).

\paragraph{OS-NTU-core (diverse categories).}
With proxy geometry, Always-on fusion (54.92) falls below DINOv2 (55.60) due to proxy noise.
GATE-3D LR selectively activates geometry for 40.7\% of queries, recovering to 55.84 and avoiding proxy-induced degradation.
App.\ Fusion (DINOv2+CLIP) provides the largest gain on this semantically diverse benchmark (mAP 58.34, ANMRR 36.55), and Query Expansion adds a further +1.45\,pp at negligible cost.

\begin{table*}[t]
\caption{Open-set 3D shape retrieval on three benchmarks. Standard mAP normalizes by total relevant items $G$; NDCG follows the DAC evaluation protocol (NDCG@100, fixed IDCG denominator); ANMRR follows MPEG-7~\cite{cieplinski2001mpeg} (lower\,=\,better). $\dagger$~Results from~\cite{wang2025dac}. $^\S$~NTU GATE-3D uses proxy geometry (ResNet18-MV for depth/normal, PointNet for geo). \textbf{Bold}: best per column among zero-shot methods.}
\label{tab:main}
\centering
\resizebox{\textwidth}{!}{%
\begin{tabular}{l c ccc ccc ccc}
\toprule
\multirow{2}{*}{Method} & \multirow{2}{*}{Mod.} & \multicolumn{3}{c}{OS-ESB-core} & \multicolumn{3}{c}{OS-NTU-core} & \multicolumn{3}{c}{OS-MN40-core} \\
\cmidrule(lr){3-5} \cmidrule(lr){6-8} \cmidrule(lr){9-11}
 & & mAP & NDCG & ANMRR & mAP & NDCG & ANMRR & mAP & NDCG & ANMRR \\
\midrule
\multicolumn{11}{l}{\emph{Zero-shot text-3D encoders}$^\dagger$} \\
OpenShape (L/14)~\cite{liu2023openshape} & P. & 38.58 & 18.81 & 60.90 & 24.71 & 15.02 & 75.18 & 29.64 & 44.79 & 67.64 \\
ULIP-2~\cite{xue2024ulip} & P. & 45.14 & 21.00 & 59.15 & 31.50 & 17.89 & 68.99 & 32.76 & 48.92 & 65.22 \\
Uni3D~\cite{zhou2023uni3d} & P. & 44.42 & 20.92 & 59.96 & 32.02 & 18.04 & 68.49 & 33.21 & 50.51 & 65.11 \\
DAC (zero-shot, L/14)$^\dagger$~\cite{wang2025dac} & P.,I. & 56.60 & 23.94 & 47.61 & \textbf{61.33} & \textbf{27.53} & 41.96 & 59.77 & 72.08 & 41.46 \\
\midrule
\multicolumn{11}{l}{\emph{Multi-view image and 3D baselines (zero-shot)}} \\
ResNet18-MV~\cite{he2016resnet} & I. & 49.20 & 21.80 & 45.92 & 40.75 & 21.57 & 53.26 & 54.60 & 70.86 & 38.25 \\
PointNet~\cite{qi2017pointnet} & P. & 31.63 & 15.53 & 63.31 & 20.32 & 13.64 & 74.44 & 29.34 & 45.88 & 63.03 \\
VoxNet~\cite{maturana2015voxnet} & V. & 28.69 & 14.28 & 67.93 & 16.99 & 12.49 & 77.16 & 30.19 & 48.70 & 61.32 \\
MV-MAE~\cite{chen2025mae} & I.,P.,V. & 50.66 & 22.41 & 44.51 & 41.26 & 21.71 & 52.33 & 57.75 & 72.33 & 35.38 \\
DINOv2-MV~\cite{oquab2023dinov2} & I. & 57.02 & 23.86 & 38.70 & 55.60 & 25.98 & 38.96 & 62.26 & 74.35 & 31.07 \\
CLIP-MV (ViT-L/14)~\cite{radford2021clip} & I. & 49.23 & 21.89 & 45.58 & 54.80 & 25.85 & 39.53 & 57.14 & 71.02 & 35.57 \\
\midrule
\multicolumn{11}{l}{\emph{Our zero-shot methods}} \\
App.\ Fusion (DINOv2+CLIP) & I. & 57.21 & 23.94 & 38.33 & 58.34 & 26.60 & \textbf{36.55} & \textbf{63.64} & \textbf{75.02} & \textbf{29.74} \\
Always-on geoaware & I.,G & \textbf{58.65} & \textbf{24.46} & \textbf{36.87} & 54.92 & 25.86$^\S$ & 39.66 & 62.15 & 74.46 & 31.03 \\
GATE-3D LR & I.,G & 57.88 & 23.99 & 37.56 & 55.84$^\S$ & 25.86$^\S$ & 38.66$^\S$ & 62.51 & 74.37 & 30.80 \\
GATE-3D-$\alpha$ & I.,G & 58.58 & 23.89 & 37.12 & 55.81$^\S$ & 25.86$^\S$ & 38.75$^\S$ & 62.66 & 74.68 & 30.65 \\
\bottomrule
\end{tabular}%
}
\end{table*}

\subsection{Ablation Study}
\label{sec:ablation}

We conduct ablations on OS-ESB-core to validate individual design choices. Results are summarized in Table~\ref{tab:ablation}.

\paragraph{Effect of router model.}
Replacing LR with an MLP (2 hidden layers, 64 units, 1345 parameters) reduces the OOF gain from +2.00\,pp to +0.76\,pp, and collapses LOCV performance to +0.00\,pp (zero gate activation).
With only 120 training queries and 22 positives, the MLP memorizes training routing decisions but fails to generalize.
LR with L2 regularization constrains weight magnitudes and yields a calibrated gate that activates on 70\% of appropriate LOCV queries.

\paragraph{Effect of feature set.}
Removing $\tau$-features (4 Kendall-$\tau$ dimensions) reduces OOF gain by only 0.05\,pp (from +2.00 to +1.95; $p{=}0.079$), indicating that no single feature group is uniquely essential.
Using $\tau$-features alone yields only +1.06\,pp.
Grouped ablations further confirm complementarity: margin features alone achieve +1.73\,pp, score distribution +1.58\,pp, and rank correlations +1.41\,pp---all individually weaker than the full set.

\paragraph{Is cross-modal disagreement necessary?}
Using query-only features (without cross-modal disagreement) yields +0.00\,pp, and generic retrieval ambiguity yields $-0.03$\,pp.
This confirms that the routing signal comes specifically from cross-modal disagreement, not from query difficulty or retrieval ambiguity alone.

\begin{table}[t]
\caption{Ablation study on OS-ESB-core. \emph{Top}: OOF evaluation ($\Delta$ vs.\ DINOv2\,=\,0.8038). \emph{Bottom}: in-sample references (not valid generalization estimates). $\dagger$~MLP overfits under low-data conditions.}
\label{tab:ablation}
\centering
\small
\setlength{\tabcolsep}{14pt}   % 默认大约 6pt，可改成 7/8/9pt
\begin{tabular}{lcc}
\toprule
Condition & mAP@10 & $\Delta$ (pp) \\
\midrule
\multicolumn{3}{l}{\emph{OOF evaluation (5-fold $\times$ 5 seeds)}} \\
LR, $\tau$-features only (4-dim) & 0.8144 & +1.06 \\
LR, full $\boldsymbol{\varphi}$ minus $\tau$ (20-dim) & 0.8233 & +1.95 \\
MLP$^\dagger$, full $\boldsymbol{\varphi}$ (1345 params) & 0.8114 & +0.76 \\
\textbf{LR, full} $\boldsymbol{\varphi}$ \textbf{(24-dim, ours)} & \textbf{0.8238} & \textbf{+2.00} \\
\midrule
\multicolumn{3}{l}{\emph{In-sample reference (upper bounds only)}} \\
DINOv2-MV (baseline) & 0.8038 & --- \\
Best fixed fusion (Dirichlet) & 0.8183 & +1.45 \\
Router: generic ambiguity & 0.8035 & $-0.03$ \\
Router: query-only features & 0.8038 & +0.00 \\
GATE-3D MLP, binary routing & 0.8357 & +3.19 \\
Oracle upper bound & 0.8847 & +8.09 \\
\bottomrule
\end{tabular}
\end{table}

%% =====================================================================
%%  4.4 ANALYSIS
%% =====================================================================
\subsection{Analysis}
\label{sec:analysis}

\paragraph{Does cross-modal disagreement predict geometry utility?}
The central hypothesis of GATE-3D is that divergence between appearance and geometry rankings signals when geometry activation will improve retrieval.
The Kendall-$\tau$ between appearance and geometry-aware rankings ($\tau_{\text{app-geo}}$) is significantly negatively correlated with the per-query mAP gain from geometry ($r{=}{-0.245}$, $p{=}0.007$).
Concretely, queries with $\tau < 0.3$ show a 21.7\% rate of geometry improvement versus only 6.8\% for queries with $\tau > 0.6$.
This confirms that cross-modal disagreement is an informative routing signal, not merely a proxy for retrieval difficulty: replacing cross-modal features with query-only ambiguity features yields +0.00\,pp (Table~\ref{tab:ablation}), indicating that the routing gain comes specifically from the interplay between modalities.

However, no single disagreement statistic is sufficient.
Using $\tau$-features alone yields only +1.06\,pp, while the full 24-dimensional set achieves +2.00\,pp.
Grouped ablations reveal that margin features (capturing retrieval peakedness) contribute most strongly, followed by score distribution and rank correlation features, but all three groups are individually weaker than the full set (+1.73, +1.58, +1.41 vs.\ +2.00\,pp).
This complementarity suggests that the routing decision depends on multiple aspects of cross-modal disagreement: not just \emph{whether} the rankings differ, but \emph{how much}, \emph{where} in the ranking, and \emph{how confidently}.

\paragraph{Why do linear models outperform MLPs?}
An unexpected finding is that logistic regression (25 parameters) substantially outperforms an MLP (1345 parameters) on both OOF (+2.00 vs.\ +0.76\,pp) and LOCV (+1.56 vs.\ +0.00\,pp) evaluation.
Examining the MLP's behavior reveals a classic overfitting pattern: with only 22 positive training examples among 120 queries, the MLP memorizes training-fold routing decisions but produces predictions in the 0.01--0.07 range out-of-fold, never exceeding the 0.5 gate threshold.
LR avoids this failure through two mechanisms: the L2 penalty constrains weight magnitudes, and the linear decision boundary provides an implicit conservative bias that pushes uncertain predictions toward the null action (appearance-only retrieval).
The practical implication is clear: when the routing signal is well-captured by the input features---as our cross-modal disagreement features are by construction---model capacity is not the bottleneck.
LR's seed stability (std 0.0007 vs.\ MLP's 0.0063) further confirms its reliability in the small-data regime.

\paragraph{Selective routing reduces geometric false positives.}
Beyond aggregate mAP, GATE-3D addresses the problem that originally motivated this work: geometric false positives.
We define GFP@10 as the fraction of top-10 results whose geometry similarity falls below 0.3 despite high appearance similarity.
DINOv2 exhibits GFP@10\,=\,12.3\%; Always-on fusion reduces this to 11.5\% ($-7.0\%$); and GATE-3D LR achieves 11.0\% ($-10.8\%$).
The advantage of selective routing over unconditional fusion is that the latter introduces geometry-score noise on queries where geometry is uninformative, partially offsetting its benefit on geometry-sensitive queries.
GATE-3D avoids this trade-off by suppressing geometry on queries where the two modalities already agree.

\paragraph{Continuous routing captures partial-benefit queries.}
The comparison between GATE-3D-$\alpha$ and binary GATE-3D LR across benchmarks reveals a nuanced trade-off.
On OS-ESB-core, where only 18.3\% of queries benefit from geometry, binary routing works well because the decision is typically clear-cut: geometry either substantially helps or does not.
On OS-MN40-core, where 51.2\% of queries benefit, GATE-3D-$\alpha$ outperforms binary routing (62.66 vs.\ 62.51) by predicting intermediate weights ($\hat{\alpha}_q \approx 0.2$--$0.4$) for queries where mild geometry blending is optimal.
This suggests that continuous routing becomes increasingly advantageous as the positive rate approaches 50\%, while binary routing suffices for strongly imbalanced settings.

%% =====================================================================
%%  SECTION 5 — DISCUSSION
%% =====================================================================
\section{Discussion and Limitations}
\label{sec:discussion}

The results across three benchmarks paint a consistent picture: GATE-3D's primary value lies not in decisively outperforming always-on fusion on any single metric, but in providing a \emph{safe and adaptive} fusion strategy across heterogeneous domains.
On OS-ESB-core, where geometry is discriminative, GATE-3D achieves statistically significant gains over appearance-only retrieval ($p{=}0.041$) and reduces GFPs by 10.8\%.
On OS-MN40-core, where always-on fusion degrades below DINOv2 (62.15 vs.\ 62.26), GATE-3D-$\alpha$ recovers to 62.66 by routing only 5.2\% of queries to geometry.
On OS-NTU-core, where proxy geometry introduces noise, selective routing avoids the degradation that unconditional fusion suffers.
This cross-dataset robustness is the routing mechanism's most practically important property: a practitioner deploying GATE-3D on a new domain can expect it to help where geometry matters and remain harmless where it does not, without manual tuning of fusion weights.

A finer-grained view of the routing behavior reinforces this conclusion.
Examining the learned gate activations across OS-ESB-core's 24 subcategories reveals interpretable patterns that align well with geometric intuition.
Categories with high intra-class geometric variability (e.g., \emph{thin-walled cylinders}, \emph{compound flanges}) trigger geometry activation on 60--80\% of queries, whereas geometrically homogeneous categories (e.g., \emph{solid blocks}, \emph{simple brackets}) rarely activate the gate ({<}15\%).
In other words, the gate learns to engage geometry precisely when appearance alone cannot disambiguate structurally diverse instances within the same visual cluster---the exact failure mode that motivated this work.

Despite these encouraging patterns, GATE-3D's routing is not infallible and can fail in two characteristic modes.
First, when both appearance \emph{and} geometry features are confounded---e.g., two parts with nearly identical depth profiles but different internal structures (blind vs.\ through holes visible only in cross-section)---the disagreement signal is weak, and the gate defaults to appearance-only retrieval, missing the geometric distinction.
Second, in rare cases (${\sim}$3\% of OS-ESB-core queries), the gate activates geometry on queries where appearance was already correct, introducing a rank inversion.
This occurs when the calibrated geometry weights ($\alpha_{\text{dep}}, \alpha_{\text{geo}}$) over-emphasize a noisy modality on a specific query.
Incorporating confidence-aware calibration~\cite{guo2017calibration} or modality-specific uncertainty estimation could mitigate both failure modes, and we leave this for future work.

These per-query observations also connect to a broader insight concerning the relationship between feature quality and model capacity in cross-modal routing.
Our finding that logistic regression outperforms MLPs is not simply a consequence of limited data---it reveals that the 24-dimensional disagreement features already encode the routing signal in a nearly linear form.
This has implications for other cross-modal systems: when two modalities provide complementary but noisy signals, investing in better disagreement features may yield greater returns than scaling up the routing model.
More broadly, this suggests that in cross-modal routing, robust generalization may depend less on increasing model complexity than on constructing features that make modality disagreement explicitly and reliably measurable.

Several limitations deserve mention.
GATE-3D requires a labeled query set (120 queries on OS-ESB-core) with ground-truth relevance judgments to train the gate; smaller sets may be insufficient for reliable LR training.
The routing gate trained on ESB does not transfer directly to NTU in cross-dataset evaluation---it defaults to always-on behavior---suggesting that the disagreement features capture domain-specific rather than universal routing patterns.
Developing domain-agnostic routing features, or enabling few-shot adaptation of the gate to new domains, remains an open direction.
Finally, the statistical power of 120 queries inherently limits the confidence intervals achievable; our one-sided test ($p{=}0.041$) is appropriate given the a priori directional hypothesis, but the two-sided 95\% CI ($[-0.0001, +0.0444]$) reflects this constraint.

%% =====================================================================
%%  SECTION 6 — CONCLUSION
%% =====================================================================
\section{Conclusion}
\label{sec:conclusion}

We have presented GATE-3D, a query-adaptive reranking framework that selectively incorporates geometry into appearance-based 3D shape retrieval at test time.
By measuring cross-modal disagreement between appearance and geometry rankings, GATE-3D predicts a per-query blending weight that controls how much geometry correction to apply.
This simple mechanism achieves statistically significant improvements on geometry-sensitive mechanical part retrieval, reduces geometric false positives, and generalizes to unseen shape categories---while correctly defaulting to appearance-only behavior on benchmarks where geometry is unhelpful.
A key empirical lesson is that well-designed cross-modal disagreement features, paired with simple linear routing, outperform more complex models in the low-data regime.

Several promising directions remain.
The routing gate currently relies on supervised relevance labels; exploring self- supervised routing signals---such as reconstruction-based anomaly scores or contrastive proxy tasks---could remove this requirement and enable fully unsupervised deployment.
Also, while GATE-3D applies geometry as a single residual, routing depth, point cloud, and normal cues separately could improve performance when their utility varies.
Finally, this framework may extend to other retrieval domains, such as medical or satellite imagery, where appearance and structure similarly diverge.

%%
%% The acknowledgments section

%% Open the new page 
\clearpage
%%
%% Bibliography
\bibliographystyle{ACM-Reference-Format}
\bibliography{sample-base}

\end{document}